**Automatic Assessment of Divergent Thinking in Chinese Language with *TransDis*:**

**A Transformer-Based Language Model Approach**

Tianchen Yang[1], Qifan Zhang[2], Zhaoyang Sun[1], and Yubo Hou[1]

[1] School of Psychological and Cognitive Sciences and Beijing Key Laboratory of

Behavior and Mental Health, Peking University

[2] School of Engineering and Applied Science, George Washington University

**Author Note**

This research was supported by a grant from the Chinese National Natural Science

Foundation (32271125) awarded to Yubo Hou. We have no conflicts of interest to disclose.

Correspondence concerning this article should be addressed to Yubo Hou, School of

Psychological and Cognitive Sciences and Beijing Key Laboratory of Behavior and Mental

Health, Peking University, Peking University, Beijing, China 100871. Email:

houyubo@pku.edu.cn

**Abstract**

Language models have been increasingly popular for automatic creativity assessment, generating semantic distances to objectively measure the quality of creative ideas. However, there is currently a lack of an automatic assessment system for evaluating creative ideas in the Chinese language. To address this gap, we developed *TransDis*, a scoring system using transformer-based language models, capable of providing valid originality (novelty) and flexibility (variety) scores for Alternative Uses Task (AUT) responses in Chinese. Study 1 demonstrated that the latent model-rated originality factor, comprised of three transformer-based models, strongly predicted human originality ratings, and the model-rated flexibility strongly correlated with human flexibility ratings as well. Criterion validity analyses indicated that model-rated originality and flexibility positively correlated to other creativity measures, demonstrating similar validity to human ratings. Study 2 & 3 showed that *TransDis* effectively distinguished participants instructed to provide creative vs. common uses (Study 2) and participants instructed to generate ideas in a flexible vs. persistent way (Study 3). Our findings suggest that *TransDis* can be a reliable and low-cost tool for measuring idea originality and flexibility in Chinese language, potentially paving the way for automatic creativity assessment in other languages. We offer an open platform to compute originality and flexibility for AUT responses in Chinese and over 50 other languages (https://osf.io/59jv2/).

*Keywords:* assessment, creativity, divergent thinking, semantic distance, transformer-based language model, originality, flexibility, natural language processing

**Introduction**

Measuring creativity reliably has been a significant challenge in creativity research. Divergent thinking (DT), the cognitive ability to generate multiple creative ideas for a specific problem (Acar & Runco, 2014), has been widely used to represent an individual's creative potential (Guilford, 1967). Various methods have been developed to evaluate the quality of creative ideas, particularly in terms of their novelty (originality) and variety (flexibility) (Guilford, 1967; Reiter-Palmon et al., 2019). For example, the most commonly used approaches are the subjective method and the empirically based method (Forthmann et al., 2020; Reiter-Palmon et al., 2019; Torrance, 1972), both of which can be applied to scoring originality and flexibility. The subjective method relies on the subjective ratings provided by multiple raters, who evaluate each response or ideational set as a whole and determine how creative these ideas are. The empirically based method assesses originality and flexibility by utilizing standardized response norms within a sample, which entails raters categorizing the responses. Greater numbers of uncommon responses and unique categories indicate higher levels of creativity.

While these methods have demonstrated practicality and reliability to some extent, three challenges remain. The first is subjectivity. Raters are not always consistent in their criteria for defining creative responses or identifying similar answers that belong to the same category. Secondly, the labor cost is high. Extensive training is required to improve consistency (Beaty & Johnson, 2021)[1]. Raters must individually score and check thousands of responses, which can be tiring and lead to reduced reliability. To enhance reliability, the most common approach is to increase the number of raters and obtain estimates of creativity scores by averaging multiple raters' ratings or through a latent variable approach. However, this can lead to an increase in labor costs and necessitate additional training for consistency. Lastly, the third challenge is sample dependency. Creativity scores obtained by the aforementioned methods are not absolute but relative to the participants in the selected sample and cannot be compared across samples (Olson et al., 2021).

---

[1] According to Beaty & Johnson (2021), raters should familiarize themselves with all facets of creativity, such as uncommonness, remoteness, and cleverness, before making judgments. It is important for them to quickly review all responses to discern commonness and uniqueness trends. The entire scale should be utilized to ensure an approximately normal distribution of scores. After completing the initial rating, revisions are encouraged to ensure accuracy.

To address these issues, researchers have started to employ language models to offer objective, automatic, and absolute scoring of creativity (Dumas & Dunbar, 2014; Prabhakaran et al., 2014). Evidence showed that language models could generate reliable and valid measures of creativity by computing semantic distances between responses and prompts as DT originality scores (Acar & Ronco, 2014; Beaty & Johnson, 2021; Dumas et al., 2021) or between responses as DT flexibility scores (Grajzel et al., 2023a; Johnson et al., 2021). However, there is currently a lack of an automatic assessment system for the novelty (originality) and variety (flexibility) of creative ideas in Chinese. Chinese vocabulary and syntactic structures have unique characteristics, such as the use of Chinese characters (rather than words) as basic units, characters with multiple meanings, and the absence of separators (e.g., spaces in English) between words. These features significantly differentiate Chinese language processing from that of English, rendering commonly used language models unsuitable for automatic creativity assessment in Chinese. This study aims to develop and validate an open-source tool for automatic creativity assessment in the Chinese language, focusing on the Alternative Uses Task (AUT). This will enable researchers to easily and objectively measure the originality and flexibility of creative ideas in commonly used creativity tasks.

**Current Research on Measuring Creativity and DT**

Creativity is often conceptualized and measured by divergent thinking (DT) (Guilford, 1967; Plucker & Makel, 2010; Reiter-Palmon et al., 2019). DT tasks typically use open-ended questions to assess an individual's ability to produce a variety of solutions (Acar & Runco, 2014; Guilford, 1950), and the responses to DT tasks often consist of longer texts, making it difficult to score objectively. The most popular DT task is the Alternative Uses Task (AUT; Guilford, 1967; Torrance, 1972). AUT responses are usually scored on three dimensions: fluency (number of responses), originality (novelty of responses), and flexibility (variety of responses). Scoring fluency is relatively simple, as it usually requires multiple raters to count the number of non-repeated relevant responses per participant (Reiter-Palmon et al., 2019). However, both the reliability (Barbot, 2018) and validity (Plucker et al., 2011) of fluency have been criticized. Moreover, a notable limitation of fluency is that it does not take into account the quality of ideas. One can generate many similar clichés but still achieve a high fluency score. Therefore, it is recommended to use an indicator of response quality in conjunction with fluency (Reiter-Palmon et al., 2019). For example, when originality is calculated by dividing it by fluency, rather than summing each response's originality ratings, it improves the discriminant validity of originality scores and helps mitigate the confounding effect of fluency (Acar et al., 2022).

Originality measures the novelty of creative ideas. The most commonly used scoring methods for originality are the subjective method (Hass et al., 2018; Silvia et al., 2008) and the empirically based method (Forthmann et al., 2020; Torrance, 1972). Subjective methods, based on Consensus Assessment Techniques (CAT; Amabile, 1983), typically involve gathering a group of raters and training them to assess the quality of creative responses. Each rater needs to rate all responses using a 5-point scale (e.g., 0 = very common, 4 = very creative; Dumas et al., 2021). Although the training provides a basic definition of what a creative response should be (e.g., unusual, distant, clever), the scoring process still relies on raters' subjective perceptions of creativity (Cseh & Jeffries, 2019). The subjective nature of CAT introduces challenges related to inter-rater reliability (Barbot, 2018). Several strategies can be employed to enhance reliability, such as providing comprehensive training to raters on the criteria and resolving disagreements during the scoring process (Amabile, 1983; Cseh & Jeffries, 2019). Another commonly used approach is to increase the number of raters and utilize the average score from multiple raters as a more refined estimate of true originality (Dumas et al., 2021). In this case, disagreements among raters on ordinal codes may actually be instructive and valuable; for instance, if half of the raters give a score of 3 and the other half give a score of 4, the average originality score of 3.5 would be considered closer to "true". However, it is worth noting that all the aforementioned methods would further increase labor costs and research time. To reduce the overall burden on raters, researchers could employ the snapshot scoring method (Silvia et al., 2009), which assesses each ideational pool as a whole. A recent study demonstrated that snapshot scoring has equivalent validity compared to laborious subjective scoring (Shaw, 2021), but it is important to note that the reliability of snapshot scoring may decrease when raters lack expertise (Hass et al., 2018).

　　　　The empirically based scoring method can also be called the frequency-based scoring method. Empirically scored originality reflects the uncommonness of responses based on the statistical frequency of each response in the study sample (Forthmann et al., 2017). Usually, raters must first aggregate all participants' responses into a response-occurrence table, which requires a lot of time to identify and combine the equivalent responses. The process of identifying equivalent responses still relies on raters' subjective judgment. Then the number of statistically uncommon responses (e.g., responses given by 5% and below of participants; Hao et al., 2017) is recorded as the originality score. The choice of the threshold varies from study to study, with past studies defining 1%, 5%, 10%, and even 20% of the sample responses as uncommon (Plucker et al., 2014). However, the frequency-based scoring method suffers from decreased accuracy when the sample size is small (Reiter-Palmon et al., 2019). Another limitation of frequency-based scoring is that instructions often encourage participants to come up with creative responses while avoiding common ones. As a result, strict adherence to the instruction may result in common responses being rarely presented and paradoxically being scored as highly original (Forthmann et al. 2017).

Flexibility scores reflect the variety of one's idea set (Guilford, 1967; Torrance, 1972). Scoring flexibility typically employs empirically based scoring, with occasional utilization of a subjective snapshot assessment method for rapid scoring (Reiter-Palmon et al., 2019). The scoring of flexibility often involves creating a category system for the responses, which is similar to the process of identifying identical responses to count their frequency during empirically-based scoring of originality. Two common approaches that rely on response categorization are used for scoring flexibility. One is counting the total number of different categories of ideas in a participant's response set (Reiter-Palmon et al., 2019). This approach requires raters to categorize each response into a conceptual category. During scoring, raters usually refer to a preexisting category system created by previous researchers on a large scale (e.g., Torrance, 1998) or create an ad hoc category system based on the given data (e.g., Yang & Wu, 2022; for a detailed procedure on creating a category system see Reiter-Palmon et al., 2019). Another variant of flexibility scoring is the number of category switches (George & Wiley, 2019; Guilford, 1967; Nusbaum & Silvia, 2011), where raters mark each response as either a conceptual switch or non-switch compared to previously generated responses. Switch scores can offer a valuable alternative for studying the shifting process during creative thinking (Preiss, 2022; Yu et al., 2019). Additionally, flexibility scoring can also employ subjective scoring, using the snapshot scoring method. In this approach, the flexibility of ideational pools relies on rater-based scoring using a Likert-type scale, ranging, for example, from 1 (not flexible at all) to 5 (very flexible) (Reiter-Palmon et al., 2019).

Researchers have also proposed another DT index called persistence, which assesses the opposite aspect of flexibility, emphasizing the depth rather than the variety of generated ideas. According to the dual pathway theory of creative thinking, there are two distinct pathways for idea generation: flexibility and persistence (De Dreu et al., 2011; Nijstad et al., 2010). The flexibility pathway involves flexible thinking, characterized by shifts among broad cognitive categories and perspectives, whereas the persistence pathway centers on systematic, in-depth exploration within limited categories. To measure persistence, researchers count the number of ideas within a few specific categories in the generation (i.e., within-category fluency; e.g., Nijstad & Stroebe, 2006). Consequently, the reliability of persistence also depends on the reliability of flexibility.

Although interrater reliability is usually high in flexibility scoring, the subjectivity is greater than in fluency scoring. Raters must subjectively determine which responses are sufficiently similar to be placed in the same category. This process can be very time-consuming when the response pool is large, as a single response might reasonably be assigned to multiple categories as the number of categories grows. The ambiguity of grouping responses complicates flexibility scoring, yet this issue is rarely addressed in the DT literature (Sung et al., 2022).

As reviewed in the previous sections, both the scoring of originality and flexibility encounter two major challenges: subjectivity and labor costs. Whether it is subjective scoring or empirically-based scoring, the scoring process involves a substantial amount of subjective judgment, including but not limited to assessing the novelty of each response, determining whether two responses belong to the same category or are identical, and evaluating the overall flexibility or originality of the entire set of responses.

 An additional issue, known as sample dependence (Silvia et al., 2008), is also a common challenge shared by both empirically based and subjective scoring methods. Both methods essentially assess individuals within a sample, using the sample as a reference frame. Regarding empirically based scoring, it is important to note that the category system is constructed based on the specific dataset being analyzed. Consequently, the category systems used for rating both flexibility and response uncommonness can vary significantly across different studies. This inconsistency results in scores that are highly sample-dependent (Forthmann et al., 2019).

Subjective scoring also involves referencing the overall response set. Although subjective scoring provides instruction to raters about what a creative response is, raters still need to develop their own definitions of creative responses based on the full list of ideas within the given sample (Reiter-Palmon et al., 2019). Therefore, the rating criteria in one study may not be the same as in another study due to differences in raters and the reference frame based on the idea list. For instance, a student's responses may not be considered outstandingly original within a leading university sample but could be evaluated as highly creative when placed in a more general sample. Since the scores of participants in different studies are not absolute, it becomes challenging to reanalyze data from various studies. This also creates difficulties for longitudinal studies and meta-analyses aiming to assess creativity across different time periods and populations.

## Automated Scoring of DT Tests Using Semantic Distance

Since Mednick's (1962) proposal of the associative theory of creativity, researchers have frequently approached creativity from the perspective of associative distance (Kenett et al., 2014; Rossmann & Fink, 2010; Kenett & Faust, 2019). Within this framework, the originality of AUT can be conceptualized as the associative distance between participants' generated ideas (e.g., to build a house) and provided prompts (e.g., bricks). After being trained on large datasets of text, language models represent words or sentences as numerical vectors in a high-dimensional space. The vector representations capture the meaning and semantic relationships between words or sentences, which enables the calculation of semantic distances between them. The theoretical pairing of associative distance and semantic distance has led researchers to employ semantic distance as a measure of DT originality: [Originality = Semantic Distance = 1 – Cosine Distance($p$,$r$)], where $p$ and $r$ represent the vectors of *prompts* and *responses*, respectively (e.g., Beaty & Johnson, 2021; Dumas et al., 2021). The semantic distance values range from 0 to 2, with higher scores indicating that the response is more distantly associated with the prompt and, hence more creative.

Although the literature on automatic scoring DT mainly focuses on the scoring of originality, recent studies started to apply semantic distance to flexibility scoring. Johnson and colleagues (2021) assessed the idea diversity as an alternative to flexibility ratings by calculating the semantic distance between participants' responses in a word association task—the corpus-based assessment of novelty and diversity (CBAND). In this task, participants were presented with a series of nouns and instructed to "think creatively" while generating words that could be creatively associated with the given noun. Sung and colleagues (2022) used the k-means clustering algorithm (MacQueen, 1967) to arrange the word vectors of participants' responses into different clusters, with each cluster representing a semantic category, and the flexibility score for a participant was determined by the total number of distinct semantic categories that a participant's responses fell into. In the aforementioned two studies, word embeddings were used to determine flexibility scores, but this method may not be suitable for AUT, where responses are typically sentences. To address this issue, Grajzel et al. (2023a) used language models to directly compute semantic distances between participants' responses in AUT and Unusual Uses Task (UUT; Torrance, 1998) as flexibility scores. Semantic distance flexibility was found to be positively correlated with human-rated flexibility and could predict the openness personality trait. Preliminary evidence indicates that language models can be used for measuring the novelty and variety of AUT responses, but their applicability to AUT in Chinese language remains to be tested. The suitability of commonly used models for processing Chinese sentences will be discussed in the following section.

**Language Models for Calculating Semantic Distance**

Latent Semantic Analysis (LSA; Landauer et al., 1998) was the earliest model used for creativity assessment. LSA reduces the dimensionality of the word-document co-occurrence matrix through Singular Value Decomposition (SVD; Golub & Reinsch, 1971), constructing a latent semantic space. Semantic distances between words can be calculated based on the vector representations of each word in this semantic space. Although LSA has been shown to provide reliable and effective measures for DT tasks (Forster & Dunbar, 2009; Dumas & Dunbar, 2014; Prabhakaran et al., 2014), it is less suitable for long-sentence responses in AUT (Forthmann et al., 2017; Harbison & Haarmann, 2014) than single-word responses (Heinen & Johnson, 2018).

This may be due to LSA's inability to take into account the context and the structure of language. Furthermore, LSA's computational complexity grows exponentially with the size of the training text, making it infeasible for large numbers of documents, resulting in a potentially less comprehensive semantic space (Sung et al., 2022). LSA is seldom used in creativity research involving non-English languages because of its heavy reliance on word co-occurrence in large text corpora and assumption of similarity in word usage patterns across languages. This assumption may not hold for non-English languages due to the differences in word order, syntax, and grammar.

To further overcome the shortcomings of LSA, Dumas and colleagues (2021) compared whether GloVe 840B (Pennington et al., 2014) and Word2Vec (Mikolov et al., 2013) would outperformed LSA models such as TASA LSA (Landauer & Dumais, 1997) and EN 100k LSA (Günther et al., 2015) in assessing the originality of AUT. The results showed that the latent factor scores of all the language models correlated with human-rated originality scores in a generally acceptable range ($r = .58 - .73$), and the language models showed similar correlations to other measures of creativity as human raters did. Beaty and Johnson (2021) also used LSA, Word2Vec, and GloVe models to generate originality scores for AUT and word association tasks. Through latent variable modeling, they found that the latent semantic distance originality factor, composed of the common variance of multiple language model ratings, could reliably predict human-rated originality and other creativity-related external validity criteria. The aforementioned research validated the effectiveness of language models in English DT assessment including AUT. However, research validating the application of language models to Chinese creativity scoring is scarce, and the existing studies have primarily focused on word-word distances rather than distances between sentences (Sung et al., 2022).[2]

Word2Vec and GloVe represent the semantic meaning of words based on their context (Harris, 1954), but this semantic representation is static. This means the models have identical representations for homonyms, regardless of the context in which they appear. Therefore, they cannot solve the problem of polysemy. Polysemy is a linguistic phenomenon present in all languages, including both English and Chinese. Statistical data reveals that over 40% of English words have multiple meanings (Traxler, 2011), while approximately 23% of entries in The Contemporary Chinese Dictionary (5th Edition) exhibit polysemy (Wang, 2009). For instance, the word *报销* has the same vector in the two sentences 那旧灯泡*报销*了 "That old light bulb is broken" and 我去财务处*报销*差旅费 "I went to the financial office to claim my travel expenses." The word *报销* means "(be) broken" in the first sentence and means "claim expenses" in the second. The aforementioned models cannot distinguish the semantic difference of the same word in different contexts. In Chinese, polysemy goes beyond words and extends to individual Chinese characters. Unlike alphabetic languages, where words are composed of letters and polysemy typically occurs at the word level, each Chinese character carries a specific meaning. Moreover, when combined with one or more characters, the meaning of a Chinese character can change, rendering language processing highly reliant on context (Bessmertny et al., 2020).

---

[2] Sung and colleagues (2022) constructed and validated a computerized creativity assessing system based on a figure association task, in which the originality and flexibility scores were calculated based on a Word2Vec language model. This scoring system indirectly generated the originality score by calculating the semantic distances between the noun extracted from the response and noncreative benchmark responses, which require word segmentation of the multi-word Chinese responses. Therefore, it cannot calculate the semantic distance between the response and the AUT prompt directly as most English AUT scoring systems do.

Statistical data shows that the average number of meanings for single-character words in Chinese is 1.5 times that of two-character words (Wang, 2009). Consequently, the commonly used static language models may not be suitable for evaluating creativity in the Chinese language.

To address these limitations, we sought to use transformer-based language models to represent the AUT responses in Chinese. Transformer-based language models, such as Google's Bidirectional Encoder Representations from Transformers (BERT; Devlin et al., 2019) and the Generative Pre-Trained Transformer 3 (GPT-3; Brown et al., 2020), use multiple layers of information to produce word embeddings that are *dependent* on context. Transformer-based language models use a self-attention mechanism to learn contextual relationships between words in a sentence or sequence. This allows each word's representation to influence the others in a sentence so that the model can better understand the meaning and context of a sentence.

Transformer-based models have several advantages in training on Chinese language data. Firstly, they can better capture long-range dependencies and syntactic structures in text. Secondly, transformer-based models can handle polysemy and homonymy in Chinese characters more effectively by considering the context in which the characters appear. Thirdly, transformer-based models can learn contextualized representations of words and phrases, which can capture variations in meaning across different contexts. Finally, transformer-based models can be pre-trained on large amounts of unlabeled data and then fine-tuned on specific downstream tasks, resulting in improved performance (e.g., SBERT is an extension of BERT that is capable of generating sentence embeddings for multiple languages through further training; Reimers & Gurevych, 2020).

In support of the use of Transformer-based models for scoring AUT, a recent study by Organisciak et al. (2023) demonstrated that fine-tuned generative large-language models based on Transformers, such as fine-tuned GPT-3, outperform non-Transformer-based models (Beaty et al., 2021; Dumas et al., 2021). However, it is important to note that this study did not employ a semantic distance-based method; instead, it had the system directly generate scores for AUT uses (e.g., Input = "autscore question: What is a surprising use for a BOOK response: relay race marker", Output = 5). Johnson and colleagues (2022) found that BERT is better suited than LSA and Word2Vec models for calculating the semantic distances between words in a five-sentence creative story written by participants, indicating its potential suitability for sentence-level creativity assessment. Therefore, the outstanding performance of Transformer-based models in English shows promise for assessing originality and flexibility in Chinese.

**Present Research**

Given that transformer-based models generate context-dependent embeddings for sentences and words, they may be highly suitable for the automatic assessment of DT tasks in Chinese. In the current research, four transformer-based models and one Word2Vec model, all of which are open-source language models for Chinese, were selected and tested to determine their suitability for measuring the novelty of responses in the AUT (details about these models can be found in the Method section of Study 1).

The research aims to accomplish three main goals across three studies: First, to identify the models with the best predictive performance for human originality ratings, in order to construct the *TransDis* automatic assessment system (Study 1). Secondly, to compare the correlation of *TransDis*-generated originality and flexibility scores with human ratings (Study 1). Thirdly, to examine the criterion validity (Study 1) and Known-Group validity (Study 2: whether the model-rated originality could discriminate between creative instruction group vs. common instruction group; Study3: whether the model-rated flexibility could discriminate between flexible instruction group and persistent response group) of the *TransDis* scores. Our overarching goal in constructing and validating *TransDis* is to provide creativity researchers with objective and automated scoring tools for DT responses in Chinese.

## Study 1

In Study 1, we first aimed to compare the predictive performance of semantic distance originality scores of several language models on human originality ratings. Higher correlations between these scores and human ratings indicated better performance. Our goal was to identify the best-performing models for constructing the *TransDis* system.

Next, we aimed to validate the *TransDis* system by two criteria: (a) whether the model-rated originality and flexibility strongly correlated with human ratings; and (b) whether the model-rated originality and flexibility exhibited a positive correlation with other creativity measures, similar to human ratings.

For the evaluation of criterion (a), we combined multiple ratings from different models into a latent variable that extracted common measurement variance across multiple models and tested whether the latent variable can approximate human originality and flexibility ratings, as Beaty and Johnson (2021) suggested.

Regarding criterion (b), we examined how the latent factor scores of *TransDis*-generated originality and flexibility relate to several creativity-related measures, including self-ratings, personality traits, and cognitive factors. Prior research has shown that self-ratings on creativity and creative self-concept positively correlated with both human and model creativity ratings (e.g., Beaty & Johnson, 2021; Yu et al., 2023). Therefore, we included the self-rated everyday creativity subscale from the Kaufman Domains of Creativity Scale (Kaufman, 2012) and the Short Scale of Creative Self (Karwowski, 2012) as external validity criteria. As for personality, many studies have found that openness is particularly correlated with creative outcomes like DT originality (Beaty & Johnson, 2021; Grajzel et al., 2023b; Kandler et al., 2016), we thus included the 8-item Openness subscale from the Big Five Inventory (John et al., 1991; John et al., 2008). Furthermore, previous work has demonstrated a positive correlation between fluid intelligence and DT originality (Benedek et al., 2012, 2014; Nusbaum et al., 2014). Additionally, intelligence and creativity have been found to share a common cognitive and neural basis (Frith et al., 2021; Jauk et al., 2014). We thus included a short form of the Raven Advanced Progressive Matrices Test (Arthur & Day, 1994) to measure fluid intelligence. We expected the *TransDis*-generated scores to positively correlate with these external measures.

**Method**

The data that support the findings of this study are openly available at https://osf.io/59jv2/.

***Participants***

This study included 350 university students (239 females; 68.3%) in China. The mean age of participants was 21.29 (*SD* = 3.00). Participants were compensated 15 Chinese yuan for their participation. Participants were all native Chinese speakers.

***Procedure***

Participants completed a series of tasks and questionnaires that measured different aspects of creative potential and creativity-related individual traits. Initially, participants completed four trials of AUT, followed by creativity-related questionnaires that included everyday creativity, creative self-efficacy, creative self-identity, and openness to experience. Lastly, they completed the fluid intelligence test. All the questionnaires and assessments were administered in a laboratory setting and conducted on a computer via www.credamo.com, a reliable Chinese data-collection platform similar to Qualtrics Online Sample.

**Alternative Uses Tasks.** Participants were assigned to work on four AUT items (bedsheet, chopsticks, slippers, and toothbrush, 2 minutes per item). In line with DT literature, the AUT prompts should be familiar to participants (Acar & Runco, 2019). To ensure this, we recruited 30 Chinese participants to rate their familiarity with 18 prompts (selected based on past DT literature) on a 7-point scale (1 = not at all, 7 = very much). The four prompts with the highest familiarity scores were chosen as the final AUT prompts (familiarity > 6.73).

In line with the prior work on automatic creativity assessment (Beaty & Johnson, 2021; Dumas et al., 2021), participants were instructed to "think creatively" when generating uses for the prompts. The instructions emphasized quality over quantity. Responses were scored for originality using the subjective scoring method (Beaty & Johnson, 2021; Benedek et al., 2014; Silvia et al., 2008). Three raters scored 6423 responses on a 5-point scale (0 = *not at all creative*, 4 = *very creative*). They were instructed that the rating standard of originality is associated with three facets: uncommonness, remoteness, and cleverness, and responses with high originality typically excel in these three aspects compared to ordinary responses (Wilson et al., 1953; Beaty et al., 2021).

Originality for each participant was finally scored using the top-scoring method (Benedek et al., 2013, 2014; Silvia et al., 2008), which could avoid the confounding effect with the fluency score. For each AUT item, the originality score reflected the average creativity rating of those three ideas that had received the highest ratings from the rater. Interrater reliability for the four AUT prompts (bedsheet, chopsticks, slippers, and toothbrush) was from fair to good (*ICC2k*

= .75, .62, .57, .74)[3]. We chose the top-3 scores over other numbers of top ideas because, as previous research showed (Benedek et al., 2013), a top-3 originality score for 2 minutes time-on-task showed the highest correlation with openness. As the correlations between originality and openness were also high when using top-2 scoring, we included the top-2 scoring results in the supplementary materials for comparison (https://osf.io/59jv2/).

The flexibility score was determined by the category switches (the first response counts as a first switch). Three raters counted the number of category switches within the generated responses for each prompt. Raters were trained to identify feature differences that resulted in functional differences in AUT (Reiter-Palmon et al., 2019). For example, saying that "a toothbrush can be used to clean a cup" is not significantly different from "used to wash shoes". However, "using a toothbrush to scratch my back" is clearly distinct from those two uses because the bristles serve a completely different function. Interrater reliability for the four AUT items was good ($ICC2k$category-switch= .84, .97, .96, .90). We also calculated the number of Chinese characters in each response as the elaboration score.

**Everyday creativity.** Self-rated creativity was measured using the everyday creativity subscale from the Kaufman Domains of Creativity Scale (Kaufman, 2012), which consists of 11 items. Everyday creativity measures how well people perform creatively in everyday life situations, including interpersonal and intrapersonal creativity (e.g., teaching someone how to do something, and understanding how to make myself happy). In this study, the 11-item everyday creativity subscale achieved an internal consistency of α = .82.

---

[3] An averaged random Intraclass correlation coefficient (*ICC2k*) using an absolute agreement definition was calculated. The two items with lower *ICC*s were later left out of the analysis.

**Creative self-efficacy and creative self-identity.** Creative self was measured by the Short Scale of Creative Self (Karwowski, 2012), which consists of two subscales, creative self-efficacy and creative self-identity. Creative self-efficacy refers to an individual's set of beliefs that she or he is able to solve problems requiring creative thinking and to function creatively. Creative self-identity subscale measures the extent to which people view creativity as a defining feature of the self-concept. In this study, the creative self-efficacy subscale achieved an internal consistency of $\alpha = .88$, and the creative self-identity subscale achieved an internal consistency of $\alpha = .86$. The whole Creative Self Scale achieved an internal consistency of $\alpha = .92$.

**Openness to experience.** The 8-item Openness subscale of the Big Five Inventory (John et al., 1991; John et al., 2008) was used to measure personality traits associated with creativity. In this study, the Openness subscale achieved an internal consistency of $\alpha = .86$.

**Fluid intelligence.** Fluid intelligence was measured by a short form of the Raven Advanced Progressive Matrices Test (Arthur & Day, 1994), which consists of 12 items. The tests consist of a series of homogeneous, progressively more difficult items that require the examinee to choose which piece (from eight options) best completes a pattern series presented across three rows of designs. The test was scored by summing the number of problems correctly solved. The present study aims to determine whether automated DT ratings similarly correlate with fluid intelligence and the other aforementioned creativity-related measures.

*Scoring with Language Models*

In this study, we evaluated the performance of five language models (Word2Vec[4], BERT[5], SBERT_mpnet[6], SBERT_MiniLM[7], SimCSE[8]) in assessing originality and flexibility (for brief model information and comparison, see Table 1).

At a response level, each model can generate originality scores for AUT by calculating the semantic distance between responses and prompts: [Semantic Distance = 1 – Cosine Distance($p,r$)] (see Fig. 1). Then at a subject level, the model-rated originality, in line with the human-rated originality, was scored using the top-scoring method: for each AUT item, the model-rated originality score was the average of the three highest-rated responses by the model.

For flexibility scoring, the model can generate flexibility scores for AUT by calculating the semantic distance between all adjacent pairs of responses (see Fig. 1). To align with the conventional summation scoring used in human-rated DT flexibility, we calculated the subject-level model-rated flexibility score by summing the semantic distances of all adjacent pairs of responses. If the participant generates only one response, the flexibility score will be zero. Our approach to flexibility scoring closely resembles the human-rated category switches (George & Wiley, 2019; Nusbaum & Silvia, 2011). However, it extends beyond, as it not only considers the number of switches but also incorporates the associative distance associated with each switch.

Code for generating originality and flexibility scores can be found at https://huggingface.co/spaces/firefighter/TransDis-CreativityAutoAssessment/tree/main. From

---

[4] Word2Vec *fastText-chinese* (Bojanowski et al., 2016): https://fasttext.cc/docs/en/pretrained-vectors.html
[5] BERT *bert-base-chinese* (Devlin et al., 2018): https://huggingface.co/bert-base-chinese
[6] SBERT *paraphrase-multilingual-mpnet-base-v2* (Reimers & Gurevych, 2020): https://huggingface.co/sentence-transformers/paraphrase-multilingual-mpnet-base-v2
[7] SBERT *paraphrase-multilingual-MiniLM-L12-v2* (Reimers & Gurevych, 2020): https://huggingface.co/sentence-transformers/paraphrase-multilingual-MiniLM-L12-v2
[8] SimCSE *simcse-chinese-roberta-wwm-ext* (Gao et al., 2021): https://huggingface.co/cyclone/simcse-chinese-roberta-wwm-ext

the following models, we would select the best-performing ones to construct the final *TransDis* system:

**Word2Vec.** Word2Vec (Mikolov et al., 2013), as a pre-training technique, represents the semantics of words based on static contextual information. It optimizes the model structure and training techniques to enable unsupervised training based on a large-scale corpus, thus overcoming the corpus-dependent problem of LSA models. In this study, the Chinese pre-trained word vectors we used were trained on Wikipedia dumps[9] (Bojanowski et al., 2016). Stop words (e.g., 但, 的) were removed as previous research suggested (Dumas et al., 2021). The stop word list was based on four commonly used Chinese stop word lists (https://github.com/goto456/stopwords), which contain 2317 words in total. We did not remove stop words when using the other four Transformer-based models (BERT, SBERT_mpnet, SBERT_MiniLM, SimCSE) because the predictive performance remained nearly unchanged (see Supplementary Table 1).

The word vectors have a dimension of 300. Sentence vectors were computed using a mean-pooling strategy (i.e., to create the sentence embedding by taking the mean of the word embeddings in the sentence).

**BERT.** BERT (Bidirectional Encoder Representations from Transformers; Devlin et al., 2018) is a language model that uses a stack of Transformers to capture deep and bi-directional information between words in a sentence. To capture the information of context in semantic processing, BERT employs a masked language model (MLM) pre-training method, where tokens (the basic meaningful units, words, or Chinese characters) are randomly masked, and the model learns to predict the masked word based on its context. The MLM objective enables deep bidirectional Transformer pre-training and facilitates the integration of left and right contexts.

---

[9] https://dumps.wikimedia.org

Researchers have employed BERT to generate sentence embeddings. The most widely adopted methods involve averaging the BERT output layer (known as mean-pooling) or using the first token of the output (referred to as CLS pooling). However, this common practice often results in rather bad sentence embeddings, often worse than averaging GloVe embeddings (Reimers & Gurevych, 2019). One possible reason is that BERT's word representations tend to cluster in a narrow cone of the vector space, rather than being uniform in all directions (Ethayarajh, 2019). While BERT is not perfect for evaluating sentence semantic distances, we include it for comparison because the subsequent language models we employ are all optimized and improved based on the BERT method. The BERT model used in this study, bert-base-chinese, was trained on Chinese Wikipedia. The embeddings have a dimension of 768. Sentence vectors were computed using a mean-pooling strategy[10].

**SBERT.** SBERT (Reimers & Gurevych, 2019), or Sentence-BERT, is a modification of BERT model. It fine-tunes the semantic space of BERT using siamese and triplet networks (Schroff et al., 2015) in a way that maximizes the similarity of semantically similar sentences and minimizes the similarity of dissimilar sentences. This allows SBERT to produce sentence vectors that are specifically optimized for cosine similarity comparisons. The labeled training data consisted of approximately 1 million sentence pairs, which were from the SNLI (Bowman et al., 2015) and the Multi-Genre NLI (Williams et al., 2018) datasets.

---

[10] The semantic distance calculated from the [CLS] embeddings of BERT showed nearly zero correlations with human ratings. Therefore, we only included mean-pooling strategy for BERT.

Due to its suitability for computing sentence-level semantic similarity, SBERT may be particularly appropriate for measuring the originality and flexibility of responses (which are typically in the form of sentences) in AUT. The current study used two multi-lingual SBERT models, mpnet and MiniLM, with different dimensions of sentence embeddings, 768 and 384, respectively. Both models were developed by training a new system to map translated sentences to the same location in a vector space as the original (monolingual) model maps the original sentences (Reimers & Gurevych, 2020). Both models employed the mean-pooling strategy as suggested by Reimers and Gurevych (2019).

**SimCSE.** SimCSE (Simple Contrastive Learning) is a novel method for fine-tuning language models to create sentence embeddings (Gao et al., 2021), which is similar to SBERT. However, unlike SBERT which uses labeled data (where similar sentences are labeled as positive pairs) to fine-tune, SimCSE uses an unsupervised approach to obtain positive pairs. Specifically, SimCSE feeds a given sentence into BERT twice, each time randomly zeroing out a fraction of the input, resulting in two slightly different embeddings for the same sentence. These two embeddings are considered positive pairs, while negative pairs are formed by pairing the sentence with all other sentences in the mini-batch. The SimCSE model used in the current study is an open-source model fine-tuned based on Chinese RoBERTa[11] (an optimized training of BERT; Cui et al., 2021), using the SimCSE method introduced by Gao and colleagues (2021).

---

[11] https://huggingface.co/hfl/chinese-roberta-wwm-ext

The embeddings have a dimension of 768. Sentence vectors were computed using the CLS pooling strategy.

### Analytic Approach

Human-rated originality, human-rated flexibility, model-rated originality, model-rated flexibility, and other external validity criteria (including everyday creativity, creative self-concept, and openness), were all modeled as indicators of their respective latent variables at a subject level using Bayesian estimation in Mplus 8. Bayesian estimation was employed in this study because it offers a more flexible analytic approach to overcome the highly restrictive features commonly applied within confirmatory factor analysis (CFA) using maximum likelihood estimation, in which cross-loadings and residual correlations are fixed at zero. In Bayesian structural equation modeling, researchers can model uncertainty in their specifications by replacing exact zero parameters with approximate zeros (i.e., zero mean, small variance), so that the model can better reflect substantive theories (Muthén and Asparouhov, 2012). Particularly in this study, we should expect cross-loadings between latent factors and observed variables in both human ratings and semantic distance scores. To avoid imposing restrictive assumptions, it is recommended to use Bayesian estimation, which better reflects substantive theories.

In all CFA modeling, the factor variances were fixed to 1, and the Bayesian iterations were set at 50,000. The posterior predictive p-value (PPp) and the 95% confidence interval for the difference in the observed and replicated $\chi 2$ values are used to assess model fit. A good fitting model is indicated when PPp values are around .50, and the 95% confidence interval values center on zero (Muthén and Asparouhov, 2012). It should be noted that in some cases, directly averaging the scores of the three models might be more appropriate. For instance, when the sample size is small ($N < 100$), researchers should carefully consider whether to use latent variables based on the accuracy of the prior information (Smid et al., 2020).

**Results**

***Identifying Suitable Models and Prompts for Automatic Assessment***

To identify the most suitable language models and AUT prompts for the final *TransDis* system, an initial set of analyses compared the performance of different language models in predicting human originality ratings across multiple AUT prompts. The originality score generated by the language models depends on the semantic distance between the prompt word and the response. The ability of a model to understand the semantics of various prompt words may vary due to differences in training corpora and methodologies. Therefore, we employed an empirical approach and chose the suitable models based on their predictive performance.

Table 2 presents zero-order correlations between semantic distance models and human originality ratings across multiple AUT prompts. Among the five language models examined, Word2Vec demonstrated the lowest predictive performance across all four prompts, with correlations ranging from .09 to .35, and was therefore excluded from the final scoring system. To build an effective assessment system, we expected the correlations to reach a moderate level ($r > .30$), thus chopsticks and slippers were excluded from the final prompts. Although BERT semantic distance positively and significantly correlated with human originality ratings in AUT bedsheet and AUT toothbrush, its correlation with rater 1 in AUT toothbrush was lower than .3 ($r = .21$). Therefore, we chose SBERT_mpnet, SBERT_MiniLM, and SimCSE as the semantic distance models, and bedsheet and toothbrush as the AUT prompts for the final *TransDis* system[12].

### Predicting Human Originality Ratings with TransDis

To examine the predictive performance of *TransDis*, we first conducted a Bayesian CFA to assess the latent correlations between the semantic distance originality factor composed of three Transformer-based language models (SBERT_mpnet, SBERT_MiniLM and SimCSE) and human-rated originality factor on the two AUT items (bedsheet and toothbrush) (see Fig. 2). We set the prior mean for major factor loadings to .70 based on the report by Beaty and Johnson (2021). The prior variance for major factor loadings was set to 25 as a large prior variance allows free estimation of the range of loading. For cross-loadings, the prior mean was set to zero and the prior variance was set to .01, as recommended by Muthén and Asparouhov (2012). For the correlated residuals, we specified an inverse-Wishart prior distribution IW (0, degrees-of-freedom parameter d = p+6), corresponding to 95% small residual covariance range of −0.2 to +0.2 (Muthén and Asparouhov, 2012). The model fits were good: PP$p$ = .51, $\chi^2$ = 95% CI [−37.41, 38.64][13]. Results showed a large correlation between the latent semantic distance originality and human-rated originality: $r$ = .93, 95% CI = [.43, .99], $p$ < .001[14]. Thus, 86.5% of the variance in human originality ratings could be explained by a latent originality factor of three transformer-based semantic distance models. Sensitivity analysis using different prior information showed that the correlation between TransDis originality and human-rated originality was robust (see Supplementary Table 3).

---

[12] The results remained consistent when applying Top-2 scoring instead of Top-3 scoring (see Supplementary Table 2). SBERT_mpnet, SBERT_MiniLM, and SimCSE consistently exhibited optimized performance in scoring the originality of AUT bedsheet and AUT toothbrush.

[13] A sensitivity analysis with different prior information is presented in Supplementary Table 1. The correlations between latent human-rated originality and latent model-rated originality remained positive.

[14] As Supplementary Figure 1 showed, the correlation between latent semantic distance originality and human-rated originality remained nearly the same when using top-2 scoring instead of top-3 scoring. $r$ = .87, $p$ < .001.

***Predicting Human Flexibility Ratings with TransDis***

To investigate the correlation between model-rated flexibility and human-rated flexibility, we modeled them individually as second-order latent variables and saved the factor scores using the SAVEDATA command for correlation calculation. The CFA model structure with one second-order latent variable (human/model-rated originality) is the same as the upper/lower half of Fig. 2 (model/human ratings as observed variables, originality for each AUT prompts as first-order factors, and overall originality as a second-order factor). This approach was chosen over modeling human and model-rated flexibility scores as two correlated latent variables within a single model because the observed variables were so closely related that the model would not be identified. Consistent with the modeling of originality, the prior mean for major factor loadings was set to 0.70, and the prior variance for major factor loadings was set to 25. For cross-loadings, we set the prior mean to zero and the prior variance to .01. For the correlated residuals, we specified an inverse-Wishart prior distribution IW (0, degrees-of-freedom parameter d = p+6).

The model fits were good: human-rated flexibility, PP$p$ = .50, $\chi^2$ = 95% CI [−20.95, 21.26]; model-rated flexibility, PP$p$ = .39, $\chi^2$ = 95% CI [−18.69, 25.89]. Results revealed a large correlation between the latent semantic distance flexibility and human-rated flexibility: $r$ = .93, 95% CI = [.43, .99], $p$ < .001. Thus, 87.0% of the variance in human flexibility ratings could be explained by a latent flexibility factor of three transformer-based semantic distance models.

***Validation with External Measures***

To examine the external validity, everyday creativity, creative self-efficacy, creative self-identity, and openness were individually modeled as a first-order factor through Bayesian CFA. Human and model-rated originality and flexibility were individually modeled as a second-order factor through Bayesian CFA. Factor scores were saved using the SAVEDATA command in Mplus 8. The fluid intelligence test was scored by summing the number of correctly solved problems. The prior information used for modeling human and model-rated originality and flexibility remained consistent with the approach detailed in the previous section. For modeling everyday creativity, creative self-efficacy, creative self-identity, and openness, the prior mean for factor loadings was set to .50, and the prior variance was set to 25.

Figure 3 presents the results of the external validation analysis. The results first indicated that regardless of the scoring method used, originality and flexibility are positively correlated ($r$ = .38 – .56). The correlation between human-rated originality and flexibility ($r$ = .51, 95% CI = [.42, .58], $p$ < .001) was almost identical to the correlation between model-rated originality and flexibility ($r$ = .52, 95% CI = [.44, .59], $p$ < .001), indicating that the discriminant validity of model-rated originality and flexibility is on par with that of human-rated originality and flexibility.

Regarding the criterion validity of originality, both human-rated originality ($r$ = .16, 95% CI = [.06, .26], $p$ = .003) and model-rated originality ($r$ = .15, 95% CI = [.05, .26], $p$ = .004) showed significant positive correlations with fluid intelligence. Model-rated originality was significantly correlated with creative self-efficacy ($r$ = .12, 95% CI = [.02, .22], $p$ = .02) and openness to experience ($r$ = .14, 95% CI = [.03, .24], $p$ = .01) and positively but non-significantly correlated with everyday creativity ($r$ = .07, 95% CI = [−.04, .26], $p$ = .22) and creative self-identity ($r$ = .08, 95% CI = [−.03, .18], $p$ = .13). However, human-rated originality did not show significant correlations with other creativity-related scales, with correlations ranging from $r = .05$ to .07, and $p$-values ranging from .21 to .35. These results suggest that model-rated originality exhibits similar and slightly better criterion validity than human-rated originality.

Regarding the criterion validity of flexibility, results showed that model-rated flexibility had similar positive correlations with creativity-related indicators as human-rated flexibility did. Model-rated flexibility was significantly correlated with everyday creativity ($r = .24$, 95% CI = [.14, .34], $p < .001$), creative self-efficacy ($r = .20$, 95% CI = [.09, .30], $p < .001$), creative self-identity ($r = .18$, 95% CI = [.07, .28], $p < .001$), fluid intelligence ($r = .11$, 95% CI = [.001, .21], $p = .048$) and openness to experience ($r = .17$, 95% CI = [.07, .27], $p = .001$). Human-rated flexibility was significantly correlated with everyday creativity($r = .22$, 95% CI = [.11, .31], $p < .001$), creative self-efficacy ($r = .18$, 95% CI = [.08, .28], $p < .001$), creative self-identity ($r = .17$, 95% CI = [.07, .27], $p = .001$), and openness to experience ($r = .17$, 95% CI = [.06, .27], $p = .001$). And human-rated flexibility was positively but non-significantly correlated with fluid intelligence ($r = .10$, 95% CI = [−.002, .21], $p = .054$). These results suggest that model-rated flexibility exhibits similar criterion validity as human-rated flexibility.

### *Semantic Distance Originality and Response Elaboration*

Previous research has shown that responses with a higher word count tend to receive higher originality scores from both human raters and semantic distance models (Beaty & Johnson, 2021; Dumas et al., 2021). In our study, we replicated these findings for Chinese responses, where human-rated originality was positively correlated with Chinese character count at the response level ($n = 3358$; $r = .19$, 95% CI = [.16, .22], $p < .001$), and the originality scores generated by three semantic distance models from *TransDis* were also positively correlated with word count (SBERT_mpnet, $r = .28$, 95% CI = [.24, .31], $p < .001$; SBERT_MiniLM, $r = .24$, 95% CI = [.21, .27], $p < .001$; SimCSE, $r = .33$, 95% CI = [.30, .36], $p < .001$). In addition, we found that the correlation between Word2Vec originality ($r = .50$, 95% CI = [.47, .52], $p < .001$)[15] or BERT originality ($r = .64$, 95% CI = [.62, .66], $p < .001$) and word count was much stronger than the correlation between human ratings and word count (Word2Vec-human: $z = 14.14$, $p < .001$; BERT-human: $z = 23.39$, $p < .001$). This finding suggests that BERT and Word2Vec may be less effective in predicting originality ratings for Chinese responses, possibly due to their tendency to overestimate the semantic differences between words and sentences with a higher word count.

## Study 2

Study 1 demonstrated that *TransDis* produced scores for originality and flexibility that closely approximated human ratings. Furthermore, the criterion validity of the automated scoring was comparable to that of human-rated scores. In Study 2, we aimed to further validate the model-rated originality by examining the Known-Group validity. Known-group validity is the degree to which an assessment is capable of discriminating between different groups that are expected to differ in predictable ways (Portney & Watkins, 1993). A meta-analysis indicated that creative instructions emphasizing the quality of DT could significantly improve originality performance in DT tasks when compared to common instruction that only emphasized quantity (Acar et al., 2020). A recent study found a great difference in originality between the creative-instruction condition and the condition that instructed participants to generate as many common and practical ideas as possible, with a Cohen's *d* of 1.45 (Chen et al., 2022). However, it remains unclear whether model-rated originality can distinguish between participants receiving different instruction. Therefore, Study 2 sought to investigate this by instructing participants to either generate creative uses or generate practical and common uses.

---

[15] Word2Vec word count was determined after the removal of stop words.

**Method**

***Participants and Study Design***

A total of 127 participants (65 females; 51.2%) in China were recruited for this study via www.credamo.com. The mean age of participants was 30.05 ($SD = 7.90$). All participants were native Chinese speakers and received a compensation of 3 Chinese yuan following the norm at www.credamo.com. They were randomly assigned to the creative condition ($n = 61$) or the common condition ($n = 66$), receiving two different AUT instructions.

In the creative condition, participants were instructed to generate creative uses for two objects (bedsheet and toothbrush). They were told to come up with creative ideas that would be considered clever, unusual, interesting, uncommon, humorous, innovative, or different (Beaty & Johnson, 2021). In contrast, participants in the common condition were instructed to generate common uses that were practical and useful, but not uncommon or unrealistic. For each AUT trial in both conditions, the participants had 3 minutes to generate uses.

A priori power analysis using G*Power (Faul et al., 2007) indicated that 51 participants in each group were the minimal requirement for a power of .80, with a medium effect size ($d = .50$), using a one-tailed independent-samples *t* test. Thus, the sample of 127 participants met the requirement.

***Scoring Originality***

In this study, two raters scored the originality for all responses on a 5-point scale (0 = not at all creative, 4 = very creative). They were instructed that the rating standard of originality is associated with three aspects: uncommonness, remoteness, and cleverness, and responses with high originality typically excel in these three aspects compared to ordinary responses (Wilson et al., 1953; Beaty et al., 2021). Participants' originality was calculated using the top-scoring method (Benedek et al., 2013, 2014; Silvia et al., 2008), consistent with Study 1. Interrater reliability for the two AUT items was satisfactory (*ICC2k* = .91, 96).

In line with Study 1, in this study, we calculated the semantic distance originality using three models (SBERT_mpnet, SBERT_MiniLM, and SimCSE) in *TransDis*. The model-rated originality was calculated by the top-scoring method, as was the human-rated originality. In this study, we computed participants' originality scores by averaging the ratings from each rater or model for every AUT prompt. Prior to averaging, the model-rated originality scores were standardized to ensure that they were aligned on a common scale.

**Results**

***Known-Group Validity of TransDis Originality***

To further validate *TransDis,* we compared the originality scores across two conditions. Replicating Forster and Dunbar's (2009) finding, results showed that the model-rated originality of participants in the creative condition (*M* = .31, *SD* = .56) was significantly greater than in the common condition (*M* = −.29, *SD* = .93), *t*(125) = 4.32, *p* < .001, Cohen's *d* = .76, 95% CI of *d*= [.41, 1.14] (see Fig. 4). We also compared the human-rated originality across two conditions and found that participants in the creative condition (*M* = 2.40, *SD* = .49) received significantly higher ratings than participants in the common condition (*M* = 1.55, *SD* = .62), *t*(125) = 8.60, *p* < .001, Cohen's *d* = 1.52, 95% CI of *d*= [1.14, 1.94]. These results indicated that originality both the *TransDis* system and human raters demonstrated satisfactory Known-Group validity in scoring originality.

## Study 3

In Study 3, we aimed to further validate the model-rated flexibility by examining the Known-Group validity. Previous studies have employed instructions to induce flexible and persistent thinking modes in DT tasks (De Dreu et al., 2011; Runco, 1991). The instructed processing mode resulted in a significant increase in flexibility scores compared to both the control group and the persistent group (De Dreu et al., 2011; Runco, 1991).

In this study, we instructed participants to generate responses for AUT through either a flexible thinking pathway or a persistent thinking pathway and examined whether model-rated flexibility can distinguish between these two groups. We hypothesized that the model-rated flexibility score would be higher in the flexible-instruction condition than in the persistent-instruction condition (see Fig. 5).

**Method**

*Participants and Study Design*

This study recruited 135 participants (82 females; 60.7%) in China via www.credamo.com. The mean age of participants was 28.37 ($SD = 8.39$). Participants were compensated 3 Chinese yuan following the norm at www.credamo.com and they were all native Chinese speakers. Participants were randomly assigned to the flexible condition ($n = 68$) or the persistent condition ($n = 67$), receiving two different AUT instructions. The instructions were adapted from previous research (Runco & Okuda, 1991; Yang & Wu, 2022; Zedelius & Schooler, 2015) and empathized the different thinking approaches to generate creative ideas. The instructions were as follows:

Flexible instruction: "*We want you to come up with as many different uses for an item as possible. Try to provide ideas from various categories. We encourage you to think flexibly, approach the problem from different angles, and focus on the diversity of uses. Research showed that this type of thinking promotes the generation of creative ideas. For example, when discussing the uses of chopsticks, you could mention "picking up vegetables," "picking up rice," and "picking up meat," but these answers all belong to the "picking up things" category. Instead, we hope you can provide a wide range of answers, such as "using chopsticks as hairpins," "throwing them like darts," "prying open a can," or "unclogging a drain." We want the uses you come up with to be unrelated and belong to different categories. Focus on diversity. This is very important to us!*"

Persistent instruction: "*We want you to explore the possible uses of an item as deeply as possible. Try to exhaust the uses within a category until you find the most creative answer. Now, we encourage you to be more persistent with your ideas, delving vertically into one aspect of the problem. Research shows that this type of thinking promotes the generation of creative ideas. For example, chopsticks have the use of "picking up things", such as "picking up vegetables,", "picking up meat," etc. If you continue to delve deeper into this aspect, you can discover more creative ideas, such as "picking up hot items," "reaching for items that are hard to reach," or "catching small insects." You do not need to frequently switch between different kinds of uses but rather exhaust a category to find the most creative one. We hope you can persist in exploring an item's uses, investigating a category vertically until you find the most creative use within it. This is very important to us!*"

A priori power analysis using G*Power (Faul et al., 2007) indicated that 51 participants in each group was the minimal requirement for a power of .80, with a medium effect size (*d*

= .50), using a one-tailed independent-samples *t* test. Thus, the sample of 135 participants met the requirement.

### Scoring Flexibility

Since the response pool in this study was too small to establish a reliable category system, we employed the snapshot scoring method to obtain a holistic flexibility rating for the participants' response set (Silvia et al., 2009; Reiter-Palmon et al., 2019). Two raters scored participants' flexibility using snapshot scoring, which requires the raters to view a participant's responses for an AUT prompt as a whole set and rate the set of responses on a 5-point scale (1 = not flexible at all, 5 = very flexible). The raters were trained to rate the responses according to the diversity of uses. Interrater reliability for the two AUT items was good (*ICC2k* = .80, 68).

In line with Study 1, in this study, we calculated the semantic distance flexibility using three models (SBERT_mpnet, SBERT_MiniLM, and SimCSE) in *TransDis*, and we computed participants' flexibility scores by averaging the ratings from each rater or model for every AUT prompt. Prior to averaging, the model-rated flexibility scores were standardized to ensure that they were aligned on a common scale.

### Results

#### Known-Group Validity of TransDis Flexibility

To further validate *TransDis,* we compared the flexibility scores across two conditions (see Fig. 6). Results showed that the model-rated flexibility of participants in the flexible condition (*M* = .22, *SD* = .93) was significantly greater than in the persistent condition (*M* = −.22, *SD* = .66), *t*(133) = 3.20, *p* = .002, Cohen's *d* = .55, 95% CI of *d*= [.20, .90]. We further compared the human-rated flexibility across two conditions, and the results showed that participants in the flexible condition (*M* = 3.15, *SD* = .61) received significantly higher ratings than participants in the persistent condition ($M = 2.35$, $SD = .74$), $t(133) = 6.83$, $p < .001$,

Cohen's $d = 1.18$, 95% CI of $d$= [.78, 1.56]. These results suggest that both *TransDis* and human raters generated flexibility scores with satisfactory Known-Group validity.

## General Discussion

This study is the first to apply Transformer-based language models from natural language

processing technology to the automated assessment of divergent thinking (DT) in Chinese

language. By comparing the performance of five models in predicting human-rated originality in

AUT, SBERT_mpnet, SBERT_MiniLM, and SimCSE were selected as the components of the

*TransDis* automatic scoring system. By conducting latent variable analysis, we discovered that

the latent originality factor generated by *TransDis* accounted for approximately 86.5% of the

variance in human originality ratings. Additionally, we introduced a new scoring method for

flexibility in AUT based on the sum semantic distance between all adjacent response pairs, and

the model-rated flexibility factor score could account for 87.0% of the variance in human

flexibility ratings. Evidence for the criterion validity of model-rated originality and flexibility

was found in Study 1. Replicating Forster and Dunbar's finding (2009), Study 2 and Study 3

indicated that originality and flexibility generated by *TransDis* were both capable of discriminating between groups of participants receiving different instructions to AUT (creative responses vs. common responses; flexible pathway vs. persistent pathway). Taken together, our findings suggest that *TransDis* provides a reliable and valid alternate to human creativity ratings for AUT in Chinese language.

### Comparing Language Models

In Study 1, we compared the predictive performance of five language models for human originality ratings to construct the final *TransDis* scoring system. Four Transformer-based models outperformed Word2Vec, consistent with our hypothesis. Furthermore, two SBERT models and SimCSE showed the greatest predictive performance and outperformed BERT. This is not surprising because both SBERT and SimCSE were fine-tuned on BERT or BERT-like language models specifically to improve the generation and comparison of sentence embeddings. In other words, SBERT and SimCSE excel at comprehending sentences, making them more suitable for evaluating AUT responses at the sentence level.

We investigated the relationship between response elaboration and the originality scores generated by different models. Consistent with prior research (Beaty & Johnson, 2021; Dumas et al., 2021), longer responses were associated with higher originality ratings by both human raters and the models. However, compared to the human raters and the three models in *TransDis*, BERT and Word2Vec significantly overestimated the originality of longer responses. This discrepancy may be due to their limitations in understanding sentence-level semantics, making BERT and Word2Vec not suitable for evaluating AUT responses in Chinese.

## Selection of AUT Prompts

The development of *TransDis* revealed that not all prompts were equally suitable for automated DT assessment. Specifically, the predictive performance of the three models in *TransDis* for the chopsticks and slippers prompts fell below the standard. This may be due to variations in semantic understanding caused by differences in the frequency of occurrence of different prompts in the training corpus or due to the low *ICC*s of the two items. Therefore, we recommend using "bedsheet" and "toothbrush" as prompts for automated DT assessment in Chinese.

However, it is worth noting that chopsticks and slippers may still become suitable for automated DT assessment if future studies yield highly reliable human ratings and demonstrate that model ratings can effectively predict human ratings on these two prompts. Additionally, it is important to note that the reliability of the measurement increases with the use of more AUT items (Beaty et al., 2022). Therefore, future research may benefit from incorporating additional AUT prompts selected based on a combination of participant familiarity and frequency of occurrence in the training corpus.

## Known-Group Validity

Study 2 and Study 3 demonstrated the Known-Group Validity of *TransDis*-generated originality and flexibility.  However, human ratings appeared to exhibit a greater discriminative effect, as evidenced by larger Cohen's *d*. One possible reason is that human raters can quickly identify features of typical common responses (e.g., using a toothbrush for brushing teeth) and promptly assign the lowest scores to answers with such features. However, there may be subtle differences among these responses that human raters overlook but are not overlooked by language models (e.g., brushing the teeth of a dog or an alligator is different from brushing teeth). This heuristic could lead to lower human-rated scores for participants in the common condition and persistent condition. To make the model ratings for typical common responses more closely aligned with human raters, one possible approach is to modify the formula for calculating semantic distance, changing from distance[prompt, response] to distance[prompt's common use, response] (e.g., Sung et al., 2022).

## Sentence-Level Semantic Distance for Different Languages

Although we initially developed TransDis for the automatic assessment of DT responses in the Chinese language, we believe this tool can also be applied to other languages, as the models we used (SBERT_mpnet and SBERT_MiniLM) are multilingual language models for over 50 languages. On our website (https://huggingface.co/spaces/firefighter/TransDis-CreativityAutoAssessment), users can select a specific model to generate originality or flexibility scores, and researchers can explore which prompts are the most suitable for their language. We expect that further research in different languages will expand our understanding of creativity and its various applications. *TransDis*, with its ability to generate sentence embeddings that effectively capture meaning, offers greater flexibility in its applications compared to word embedding-based scoring systems. Transformer-based language models have already been applied in the evaluation of creative writing. For example, Johnson and colleagues (2022) used the BERT model to generate context-dependent embeddings for words in participants' creative stories. They then calculated the semantic distance between these word embeddings as a measure of participants' ability to integrate divergent semantics. In the same vein, *TransDis* can be easily employed to compute the semantic distance between sentences, allowing evaluation of the quality of creative writing. It can also be used to assess the semantic distance between free associations in the association tasks involving sentence responses, rather than just word responses (Gray et al., 2019).

**Summary, Limitations, and Future Directions**

To summarize, our research is the first to employ Transformer-based models to generate sentence vectors for responses in AUT and to use semantic distance for automatic assessment of DT in a non-English language. Our system, *TransDis*, can produce efficient and valid indices of creative originality and flexibility for AUT, the most commonly used task for assessing DT.

To further improve the performance of TransDis, future research could explore fine-tuning the models using supervised data, allowing the models to learn the specific features of typical original and common responses. Recent research has demonstrated the robustness of fine-tuned large-language models like GPT-3 (Brown et al., 2020; Neelakantan et al., 2022) and Text-to-Text Transfer Transformer (T5; Raffel et al., 2020) in generating originality scores closely related to human ratings when directly asked to rate AUT responses, with only a small number of training examples (Organisciak et al., 2023). However, it was also observed that when using the semantic distance method, as in our study, GPT-3 and T5 embedding models achieved only an average performance of .22. While our study did not fine-tune the models with supervised data as Organisciak and colleagues (2023) did, it is worth noting that the best-performing models (SBERT and SimCSE) were fine-tuned on sentence pair data during the pre-training stage. This highlights two implications. Firstly, while Organisciak and colleagues' findings challenged the semantic distance assumption, our study shows that semantic distance can still serve as a proxy for DT but necessitates the use of appropriate models. Secondly, future research may benefit from employing models fine-tuned explicitly for AUT tasks to improve both reliability and validity. For instance, SBERT and SimCSE can be further fine-tuned through supervised data (e.g., AUT prompt-response pairs and their corresponding human ratings, to enhance their performance in evaluating creativity).

Moving forward, future research could explore the potential of using *TransDis* in different contexts, such as assessing creativity in different domains or assessing the impact of various interventions on creativity. For example, *TransDis* might be a valuable tool in educational settings, providing a quick and cost-effective method of assessing student creativity. By measuring flexibility, *TransDis* can help to identify individuals' unique thinking styles during DT and facilitate the development of more individualized instruction and targeted interventions to promote creativity in the classroom. *TransDis*' rapid scoring capabilities can also find valuable applications in assessing DT and creativity within organizational settings. It is particularly well-suited for large-scale talent screening in businesses or for identifying creative potential from vast sets of open-ended responses. For instance, prior research used employees' suggestions as an indicator of creativity in the workplace (e.g., Williams et al., 2004). By calculating semantic distances between responses, we can assess employees' DT flexibility in organizations. While tasks like suggestion tasks may not have prompts like AUT, we can calculate the average distance between responses and the most common ones as a substitute for originality scores (Sung et al., 2022). By combining sentence-level semantic distance measurements with clustering techniques to identify the most common responses (MacQueen, 1967), researchers theoretically have the potential to generate originality scores for nearly all verbal divergent thinking tasks. Future research to establish the effectiveness of such scoring methods in various domains would be highly valuable.

In addition, it is important to note that semantic distance scoring is only applicable to verbal creativity tasks and may not be suitable for figural DT tasks, such as the figural DT tests in the Torrance Tests of Creative Thinking (Torrance, 1972). This limitation also implies that semantic distance scoring systems cannot be used to assess creativity in younger children who have weaker reading and writing abilities. Future research can explore automated assessment for figural creativity tasks to facilitate research on the development of creativity. For instance, the recent multi-modal deep learning model, CLIP (Radford et al., 2021), has been designed to learn representations of visual content from natural language supervision by utilizing a vast dataset of text-image pairs. It is capable of representing both text and pictures into a high-dimensional vector space, and thus calculating the semantic distance between them. This can potentially be used in assessing creativity in figural DT tasks, as well as in other tasks that involve multi-modal inputs.

## Conclusion

The current studies provide evidence that *TransDis*, a semantic distance scoring system based on transformer-based language models, can be a useful tool for automatically assessing originality and flexibility for AUT in Chinese language. The findings highlight that *TransDis* is highly effective in calculating semantic distance for sentence-level creative responses, and its potential applicability in assessing creativity across different languages. To facilitate new discoveries across diverse disciplines and cultures, we provide an open platform for researchers to easily compute originality and flexibility scores using *TransDis* (https://huggingface.co/spaces/firefighter/TransDis-CreativityAutoAssessment).

**Table 1**

*Size of embedding models*

| Model | Model size | Number of layers | Embedding dimensionality | Context-dependent embeddings |
|---|---|---|---|---|
| 1. Word2Vec | 600M parameters | 3 | 300 | No |
| 2. BERT | 103M parameters | 12 | 768 | Yes |
| 3. SBERT_mpnet | 278M parameters | 12 | 768 | Yes |
| 4. SBERT_minilm | 118M parameters | 12 | 384 | Yes |
| 5. SimCSE | 103M parameters | 12 | 768 | Yes |

*Note.* Word2Vec = Word2Vec *fastText-chinese*, BERT = BERT *bert-base-chinese*,

SBERT_mpnet = SBERT *paraphrase-multilingual-mpnet-base-v2*, SBERT_minilm = SBERT

*paraphrase-multilingual-MiniLM-L12-v2*, SimCSE = SimCSE *simcse-chinese-roberta-wwm-ext*.

**Table 2**

*The Correlations between Semantic Distance Models and Human Ratings*

| Model | AUT bedsheet | | | AUT toothbrush | | | AUT slippers | | | AUT chopsticks | | |
|---|---|---|---|---|---|---|---|---|---|---|---|---|
| | R1 | R2 | R3 | R1 | R2 | R3 | R1 | R2 | R3 | R1 | R2 | R3 |
| 1. word | .16 | .09 | .16 | .20 | .22 | .17 | .33 | .28 | .35 | .21 | .17 | .28 |
| 2. bert | .33 | .33 | .39 | .21 | .35 | .42 | .20 | .25 | .43 | .17 | .17 | .25 |
| 3. mpnet | .47 | .36 | .48 | .36 | .46 | .60 | .16 | .29 | .35 | .33 | .33 | .40 |
| 4. minilm | .54 | .41 | .51 | .38 | .45 | .60 | .23 | .29 | .33 | .40 | .29 | .34 |
| 5. simcse | .43 | .31 | .45 | .49 | .47 | .70 | .19 | .25 | .34 | .26 | .16 | .25 |

*Note.* $N$ = 350; correlations greater than .11 are significant at $p<.05$; correlations greater than .14

are significant at $p<.01$; correlations greater than .17 are significant at $p<.001$. word = Word2Vec semantic distance, bert = BERT semantic distance, mpnet = SBERT_mpnet semantic distance,

miniLM = SBERT_MiniLM semantic distance, simcse = SimCSE semantic distance. R1-R3 =

rater1-rater3, human originality rating.

**Figure 1**

*Flexibility as the sum of semantic distances between adjacent responses. Originality as semantic distances between responses and the prompt. The arrows represent the semantic distances computed by the language model. The greater the semantic distance, the more flexible and creative the participant is.*

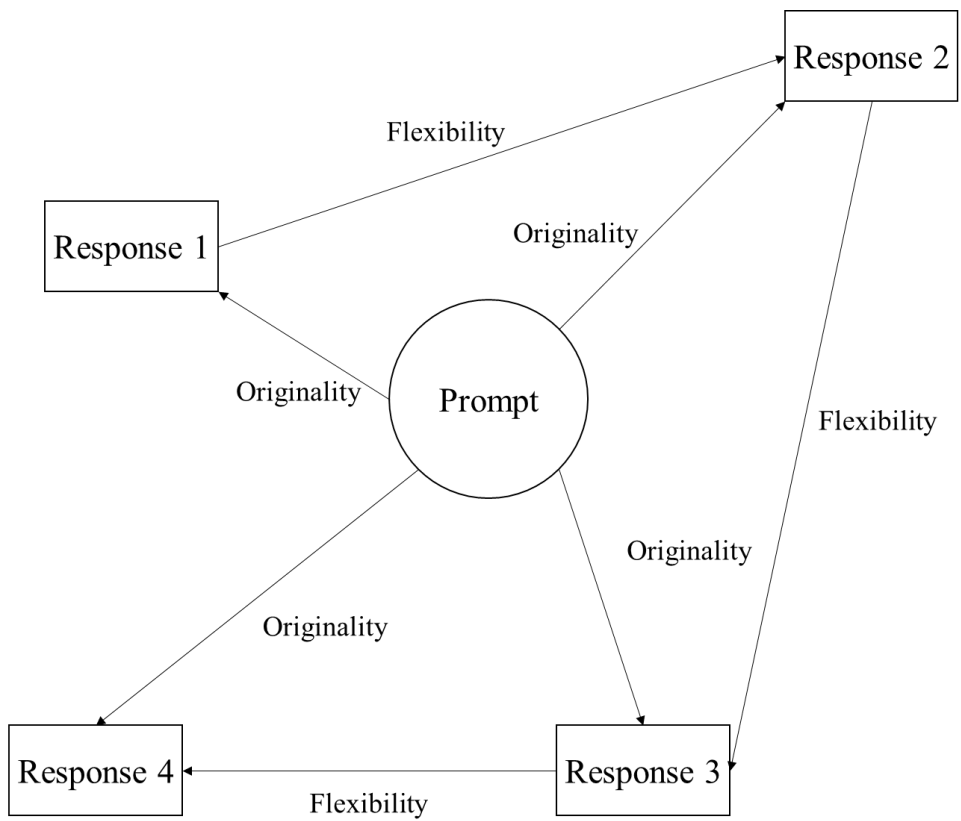

**Figure 2**

*CFA of human originality ratings and semantic distance originality for two AUT items. N = 350. hum = human rating; sem = semantic distance; b_r1-b_r3 = AUT bedsheet originality, rater 1-*

*rater 3; t_r1 –t_r3 = AUT toothbrush originality, rater1-rater3; b/t_mpnet = AUT*

*bedsheet/toothbrush, SBERT_mpnet originality; b/t_minilm = AUT bedsheet/toothbrush,*

*SBERT_MiniLM originality; b/t_simcse = AUT bedsheet/toothbrush, SimCSE originality.*

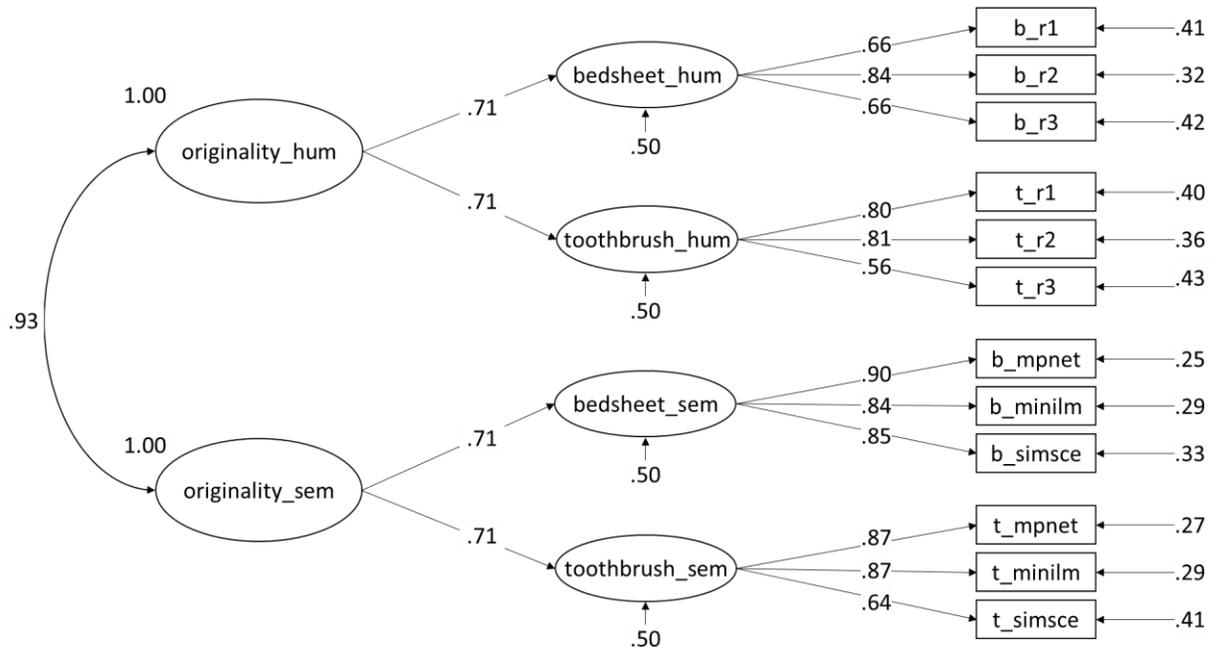

**Figure 3**

*Correlation matrix of different measures of creativity. N = 350. Hum = human rating; Sem = semantic distance; EC = everyday creativity; SE = creative self-efficacy; ID = creative self-identity; Int = fluid intelligence; Open = openness to experience. The smaller and darker the ellipse is, the larger the absolute value of the correlation coefficient is. Blue represents positive correlation, red represents negative correlation. The numbers in the lower part of the matrix are the correlation efficients. \* p < .05. \*\* p < .01. \*\*\* p < .001.*

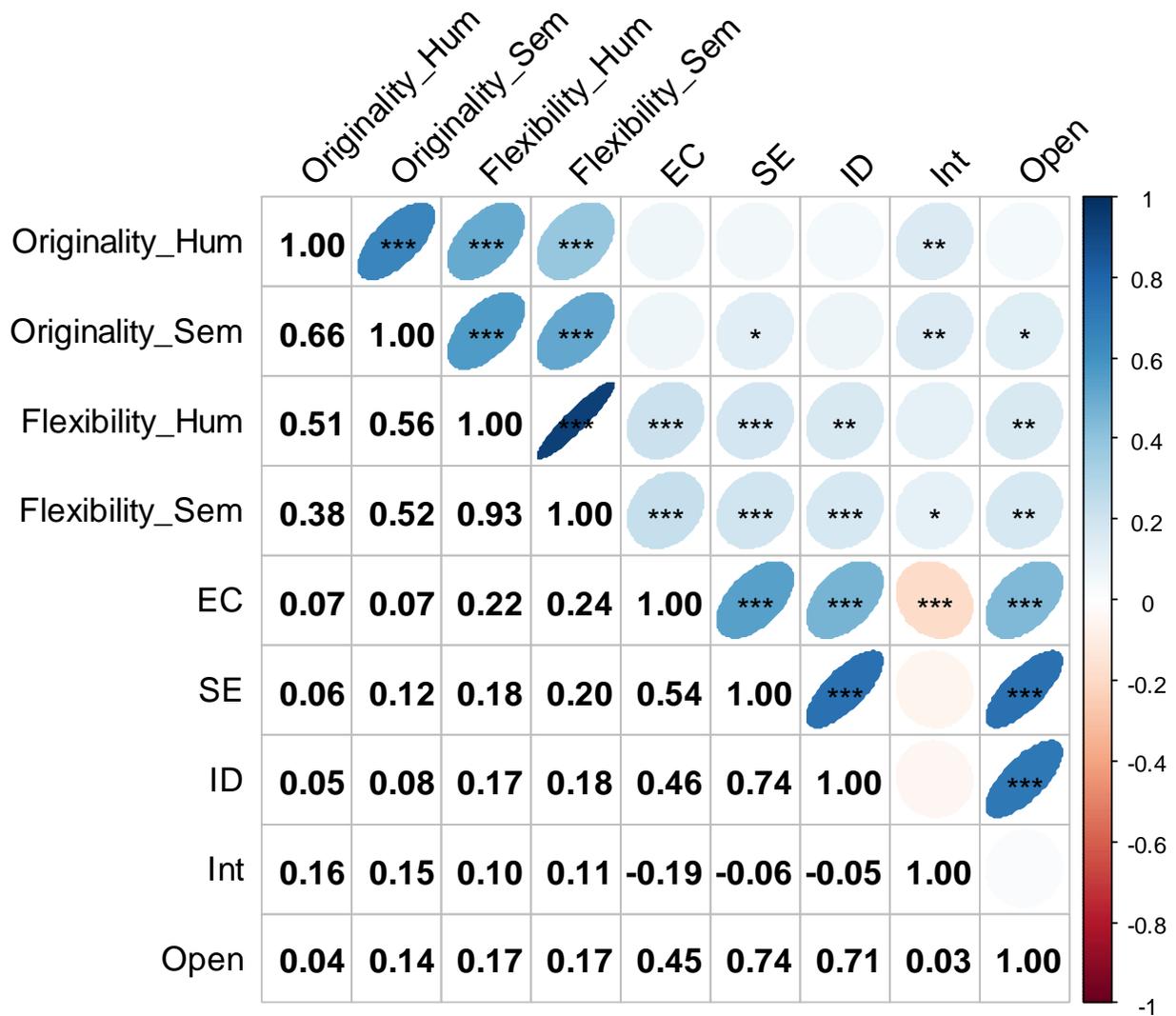

**Figure 4**

*Results from Study 2 showing both TransDis and human raters generated originality scores with satisfactory Known-Group validity. The point and short line next to the scattered points are the mean and 95% confidence intervals. The distribution of the mean difference between the two conditions is illustrated in the right section of the figure.*

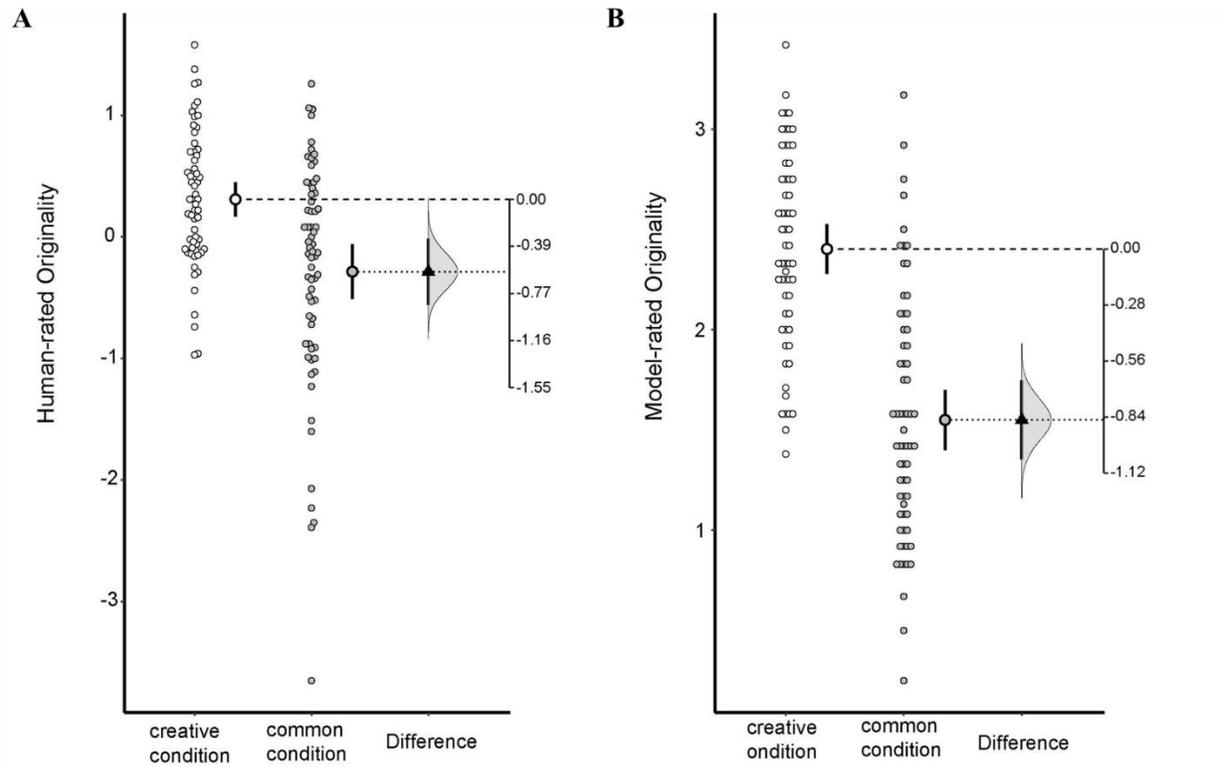

**Figure 5**

*Left: The hypothesized semantic space of responses from participants following the flexible instruction is characterized by diverse, loosely connected concepts spanning a wide range of categories. The hypothesized semantic space of responses from participants following the persistent instruction is characterized by more focused, closely related concepts within a narrow range of categories. According to our hypothesis, the semantic distance between participants' responses in the flexible-instruction condition would be greater than in the persistent-instruction condition.*

**L: Flexible Pathway**                                    **R: Persistent Pathway**

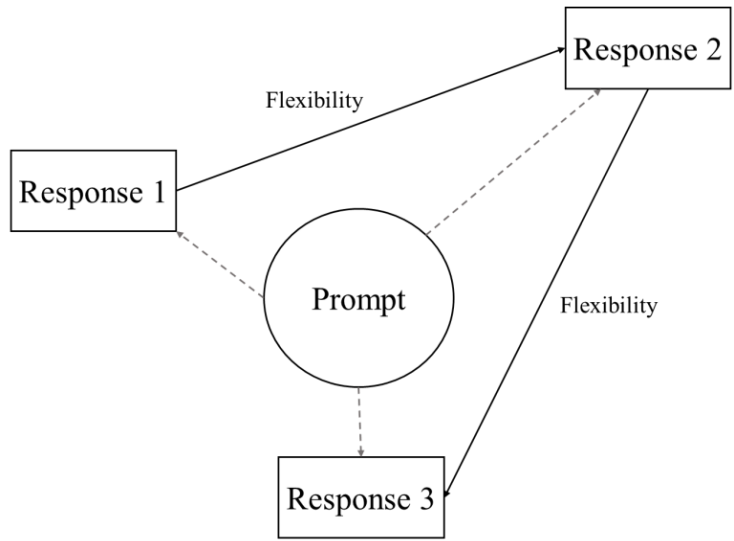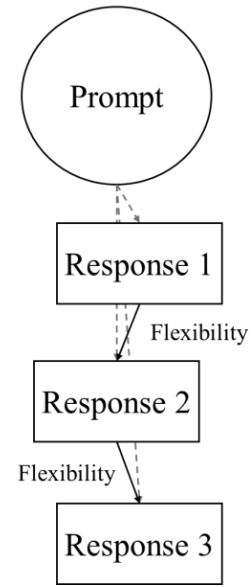

**Figure 6**

*Results from Study 3 showing both TransDis and human raters generated flexibility scores with satisfactory Known-Group validity. The point and short line next to the scattered points are the mean and 95% confidence intervals. The distribution of the mean difference between the two conditions is illustrated in the right section of the figure.*

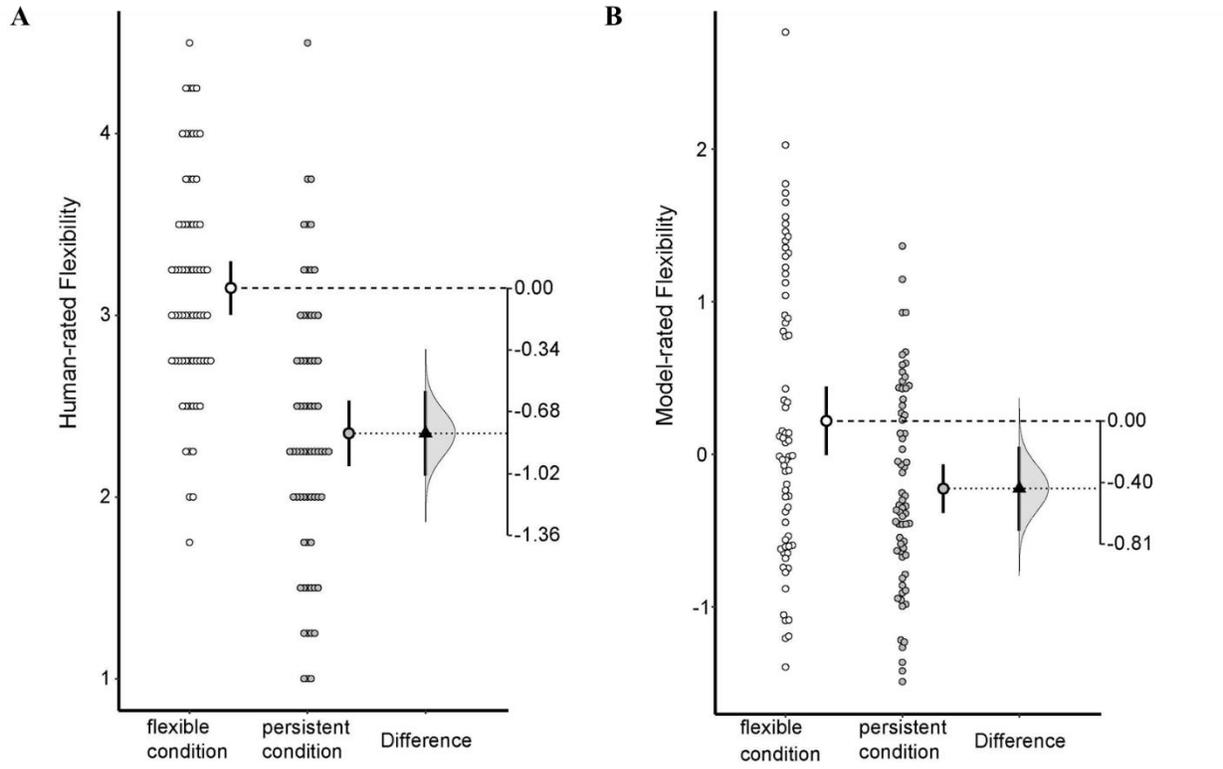